\documentclass{article}
\usepackage{spconf,amsmath,graphicx}

\usepackage{enumitem}
\setlist{nosep, leftmargin=14pt}

\usepackage{mwe} 

\usepackage{bm}
\usepackage{tabularx} 
\usepackage{url}
\usepackage{comment}

\title{JUMP: A joint multimodal registration pipeline for neuroimaging with minimal preprocessing}
\name{Adrià Casamitjana $^{1, 2}$, Juan Eugenio Iglesias $^{3, 4, 5}$, Raúl Tudela $^{6}$, Aida Niñerola-Baizán $^{6,7}$, Roser Sala-Llonch $^{2, 6, 8}$} 
\address{$^{1}$ Universitat Politècnica de Catalunya \\
        $^{2}$ Institut de Neurociències Department of Biomedicine, University of Barcelona \\
        $^{3}$ Martinos Center for Biomedical Imaging, MGH and Harvard Medical School \\
        $^{4}$ Computer Science and Artificial Intelligence Laboratory, Massachusetts Institute of  Technology \\
        $^{5}$ Centre for Medical Image Computing, University College London \\
        $^{6}$ Centro de Investigación Biomédica en Red de Bioingeniería, Biomateriales y Nanomedicina, ISCIII \\ 
        $^{7}$ Nuclear Medicine Department, Hospital Clínic\\
        $^{8}$ Institut d’Investigacions Biomèdiques August Pi i Sunyer (IDIBAPS)}

\begin{document}
%
\maketitle
\begin{abstract}
We present a pipeline for unbiased and robust multimodal registration of neuroimaging modalities with minimal preprocessing.  While typical multimodal studies need to use multiple independent processing pipelines, with diverse options and hyperparameters, we propose a single and structured framework to jointly process different image modalities. The use of state-of-the-art learning-based techniques enables fast inferences, which makes the presented method suitable for large-scale and/or multi-cohort datasets with a diverse number of modalities per session. The pipeline currently works with structural MRI, resting state fMRI and amyloid PET images. We show the predictive power of the derived biomarkers using in a case-control study and study the cross-modal relationship between different image modalities. The code can be found in \url{https://github.com/acasamitjana/JUMP}.
\end{abstract}
\section{Introduction}
\label{sec:intro}
Multimodal neuroimaging studies comprehensively describe brain variability in multiple fronts, such as brain structure, function and metabolism. Therefore, combining different sources of information is of utmost importance in neuroimaging studies aiming to study the brain, especially in dementia studies.

Typically, multimodal neuroimaging studies involve a T1w image, due to its higher resolution and better tissue contrast and another modality, such as resting-state fMRI (rs-fMRI)~\cite{ibrahim2021diagnostic}, or positron emission tomography (PET) images \cite{donohue2017association}. Studies involving more than one of these modalities \cite{biel2022combining,scherr2021effective,hampton2020resting}, usually choose the T1w image as reference and independently preprocess each modality, which could bias the result.
In addition, one of the difficulties in multimodal image studies is the large diversity of pipelines targeting each modality, making the overall preprocessing and posterior analyses tedious and time-consuming. 

This paper introduces a single multimodal preprocessing pipline for brain imaging. The core of the pipeline is the unbiased registration method followed by a minimalist approach to image preprocessing. It capitalises on previous work on contrast-agnostic registration and segmentation and it is based on our previous method for longitudinal registration, USLR \cite{casamitjana2023USRL}. The main contributions of this work are: (\emph{i})~propose a single pipeline for multimodal image processing, (\emph{ii}) develop an unbiased and joint registration method for multiple brain image modalities; and (\emph{iii}) build a minimal dedicated preprocessing step for each image modality.

\section{Material and methods}
\label{sec:method}

\subsection{Registration framework}
\label{sec:reg_method}
\begin{figure}[h]
    \centering
    \includegraphics[width=0.8\columnwidth]{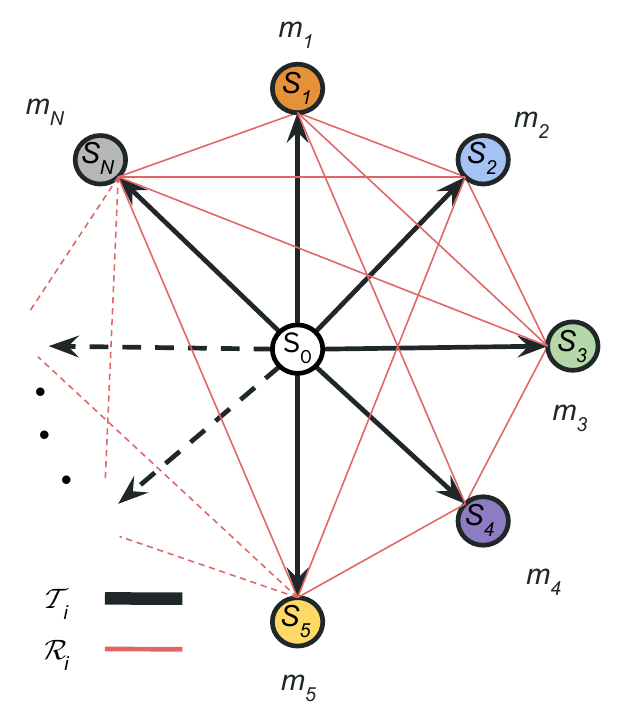}
    \caption{The ``orbit-like'' graph structure chosen. In the center, the session-specific template and the $\{m_i\}_{i=1}^N$ modalities around. In red, the dense observational graph built from pairwise registration of all modalities, while in black our choice of spanning tree.}
    \label{fig:graph}
\end{figure}

Let us consider a given imaging session with $N$ multimodal acquisitions $\{m\}^N_{n=1}$. We build a graph $\mathcal{G}$ with $N+1$ vertices where each vertex $\{S\}^N_{n=1}$ is associated with an image modality and the $S_0$ corresponds to the session template, as shown in Figure~\ref{fig:graph}.  Moreover, we define a spanning tree connecting all image modalities through the template, whose edges are associated with a set of $N$ latent transforms $\{\mathcal{T}\}_{n=1}^N$ defined from the template to the image. Finally, we consider a set of $K=N!/(2\cdot(N-2)!)$ rigid transforms $\{\mathcal{R}\}_{k=1}^K$ between each pair of images that we can compute using any registration algorithm of our choice. For fast computation of $\{\mathcal{R}\}_{k=1}^K$, we use the same rigid registration algorithm as in \cite{casamitjana2023USRL, iglesias2023ready}, which is based on an initial parcellation of each image.

We use a probabilistic model for $\mathcal{R}_k$ and $\mathcal{T}_n$, assuming that the observed registrations are conditionally independent given the latent transforms:

\begin{equation}
\label{eq:prob_model}
    p(\{\mathcal{T}_n\},  \{\mathcal{R}_k\}, \bm{\theta}) 
    = p(\{\mathcal{T}_n\}) \cdot p(\bm{\theta}) \prod_{k=1}^K p(\mathcal{R}_k | \{\mathcal{T}_n\}, \bm{\theta}),
\end{equation}
where $\bm{\theta}$ are model parameters and $p(\{\mathcal{T}_n\})$ can be seen as a regularizer. This probabilistic function can be linearised when parameterising the rigid transforms in the Lie algebra domain (also referred to as log-space) \cite{blanco2021tutorial}: 
$$\mathcal{R}_k=\exp\left[\bm{R}_k \right], \hspace{3.5mm} \mathcal{T}_k=\exp\left[\bm{T}_k\right].$$ 
In this log-domain, the inversion and composition transforms are equivalent to the negation and addition:
$$\mathcal{T}_n^{-1} = \exp{\left[-\bm{T}_n\right]}, \hspace{3.5mm} \mathcal{T}_n \circ \mathcal{T}_{n'} \approx \exp{\left[\bm{T}_n + \bm{T}^T_{n'}\right]},$$
Therefore, the generative model reads as follows:
$$
\bm{R}^j=\bm{W} \bm{T}^j+\bm{\zeta}^j.
$$
where $j=1,\ldots,6$ represents the 6 rigid parameters, $\bm{\zeta}$ is the registration error and $\bm{W}\in \mathrm{R}^{K\text{x}N}$ is a sparse matrix with two non-zero entries in each row encoding the path that each pairwise registration passes through the spanning tree, i.e., for the $k-$th registration between $m_n$ (reference) and $m_{n'}$ (target) modalities, $W_{kn}=-1$ and $W_{kn'}=1$. Thus, likelihood function can be written in terms of the log-space parameterisations as $p(\mathcal{R}_k | \{\mathcal{T}_n\}, \bm{\theta}) = p(\bm{R}_k | \{\bm{T}_n\})$, for which we use a Laplacian distribution:
\begin{equation}
\label{eq:nonrigid_laplace_likelihood}
    \bm{R}_k^j \sim \text{Laplace}\left(\bm{W}\bm{T}^j, b_M\right),
\end{equation}
where $b_M$ is the scale of the Laplace distribution and considered the same for all observations.
Finally, we regularise the optimisation by estimating a session space that lies on the centre of all modalities:
\begin{equation}
   \label{eq:laplace_prior}
    \left(\sum_{n=0}^{N-1} \bm{T}^j_n \right) \sim \text{Laplace}\left(0, b_Z\right),
\end{equation}
where $b_Z$ is the scale of the Laplace distribution. The final optimisation depends on the relationship $b_M/b_Z$ and is empirically set to 1 as in \cite{casamitjana2023USRL}. 

For more information about the inference algorithm, we refer the reader to our previous publications~\cite{casamitjana2023USRL,casamitjana2022robust}


\subsection{Pipeline}
\label{sec:pipeline}
The pipeline can be split into three different steps: first, a minimal preprocessing pipeline to prepare images for registration; second, a rigid registration step using all available modalities in each session; finally, a minimal modality-specific preprocessing step that accomodates all modalities for downstream tasks. Currently, the pipeline is set to use T1w, rs-fMRI and amyloid PET images.


\subsubsection{Initial segmentation}
In this step, we compute image segmentation using SynthSeg~\cite{billot2023synthseg}, a contast agnostic segmentation method for brain MRI. When multiple acquisitions are available (e.g., fMRI or dynamic PET) we use the mid-point image. If PET images are available, we first use SynthSR \cite{iglesias2023synthsr} to synthesise paired 1mm$^3$ MR scan prior to segmentation. 

\subsubsection{Rigid transform}
We use the method presented in Section~\ref{sec:reg_method} to compute a session-specific space for each session, where all modalities are normalised using the latent transforms. We chose the aligned T1w image and segmentation as the session-specific template due to their higher resoltuion and brain tissue contrast. This template is used in the minimal preprocessing step and for MNI registration in voxel-level population analyses.

\subsubsection{Minimal preprocessing for downstream tasks}
We use a minimalist approach to extract multimodal imaging biomarkers for downstream tasks. Structural MR images are corrected for intensity inhomogeneities using a similar approach as in \cite{van1999automated}. In resting-state fMRI, we perform motion correction across time and nuisance regression using the computed motion parameters and the average signal in white-matter tissue, brain CSF and globally in brain tissue. Finally, in PET images, we compute the SUVr values for different regions following the criteria of \cite{markiewicz2018niftypet}, using the cerebellar grey-matter as reference region. In dynamic PET imaging we initially perform motion correction and compute the average signal.

\subsection{Dataset}
In our experiments we use subjects from the ADNI dataset with at least a T1w image and one of the following modalities: (\emph{i}) resting-state fMRI and (\emph{ii}) amyloid PET regardless of the tracer. The total number of available subjects is 1247, of whom 467 have, at least, one follow-up visit. Clinical status of each session is set according to ADNI standards. For case-control analysis, we use only subjects with at least two visits that either cognitively normal (CN) or diagnosed with Alzheimer's disease (AD) through the entire follow-up period, i.e., subjects that remain stable in each diagnostic category.

\section{Experiments and results}
\label{sec:pagestyle}
We perform statistical analyses in different potential applications of the pipeline. 

\subsection{Resting-state fMRI}

\begin{figure}[h]
    \centering
    \includegraphics[width=\columnwidth]{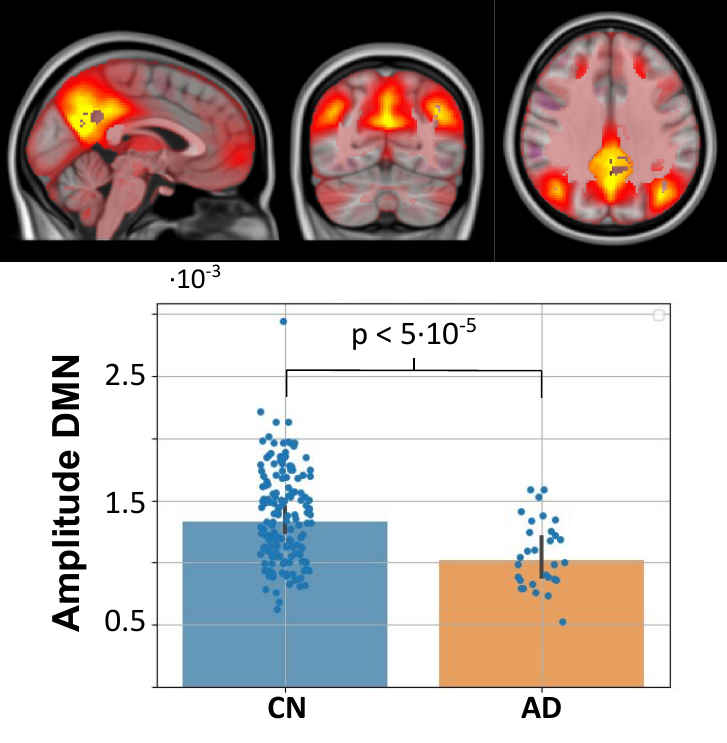}
    \caption{Top: we show the default mode network (DMN) in MNI space computed using Group-ICA over all subjects' time-series. Bottom: we compare baseline DMN amplitude for $N_{\text{CN}}=162$, $N_{\text{AD}}=30$.}
    \label{fig:rsfmri}
\end{figure}

Using CN and AD subjects with abailable resting-state fMRI we run a group-ICA  analysis with 15 components \cite{beckmann2005investigations} to detect the main networks of the sample. We clearly distinguish the motor network as the first component and the default mode network as the second component. Other networks can be found in the subsequent components similar to \cite{beckmann2005investigations}. In the first row of Figure~\ref{fig:rsfmri} we show the three orthogonal planes of the default mode network. Using the first step of a dual regression analysis \cite{nickerson2017using}, we compute the amplitude of the DMN component for each subject and test the statistical significance between groups, as shown in Figure~\ref{fig:rsfmri}.

In a data driven way, we further compute the probability of each voxel to belong to a given network which is used to parcellate the brain into 15 different ROIs, which can then be normalised to the space of each session.

\subsection{Amyloid PET}
Here, we compare SUVr between different regions and CSF biomarkers. Apart from anatomical regions, we use the DMN computed in the previous step as a region of interest (ROI). In Table~\ref{tab:suvr_ttest}, we compute group differences between CN and AD subjects for different PET and CSF biomarkers. We see that some regions (i.e., neocortex, parietal and cingulate) yield lower $p$-values than CSF amyloid-beta. Moreover, we see that the DMN appears as a potential valid biomarker for AD diagnosis.

\begin{table}[h]
    \centering
    \begin{tabular}{
    |>{\centering\arraybackslash}p{2.5cm}
    ||>{\centering\arraybackslash}p{1cm}
    >{\centering\arraybackslash}p{2cm}|}
     \hline
     \multicolumn{3}{|c|}{Biomarker comparison} \\
     \hline
     Biomarker & \multicolumn{2}{|c|}{CN vs AD}  \\
      & $t$-value & $p$-value \\
     \hline
     CSF A$\beta$        &  7.25 & 5.72·$10^{-11}$   \\
     PET neocortex & -7.54 & 1.31·$10^{-11}$    \\
     PET parietal  & -7.90 & 2.20·$10^{-12}$  \\
     PET temporal  &  -6.99 & 2.12·$10^{-10}$  \\
     PET cingulate & -7.89 & 2.31·$10^{-12}$  \\
     PET DMN & -6.76 & 6.68·$10^{-10}$  \\
     \hline
    \end{tabular}
    \caption{Effect sizes of CSF amyloid beta and amyloid SUVr values for different ROIs for CN-AD groups and stable MCI and converter MCI. Sample sizes are: $N_{\text{CN}}=68$, $N_{\text{AD}}=45$. }
    \label{tab:suvr_ttest}
\end{table}

\begin{figure}[h]
    \centering
    \includegraphics[width=\columnwidth]{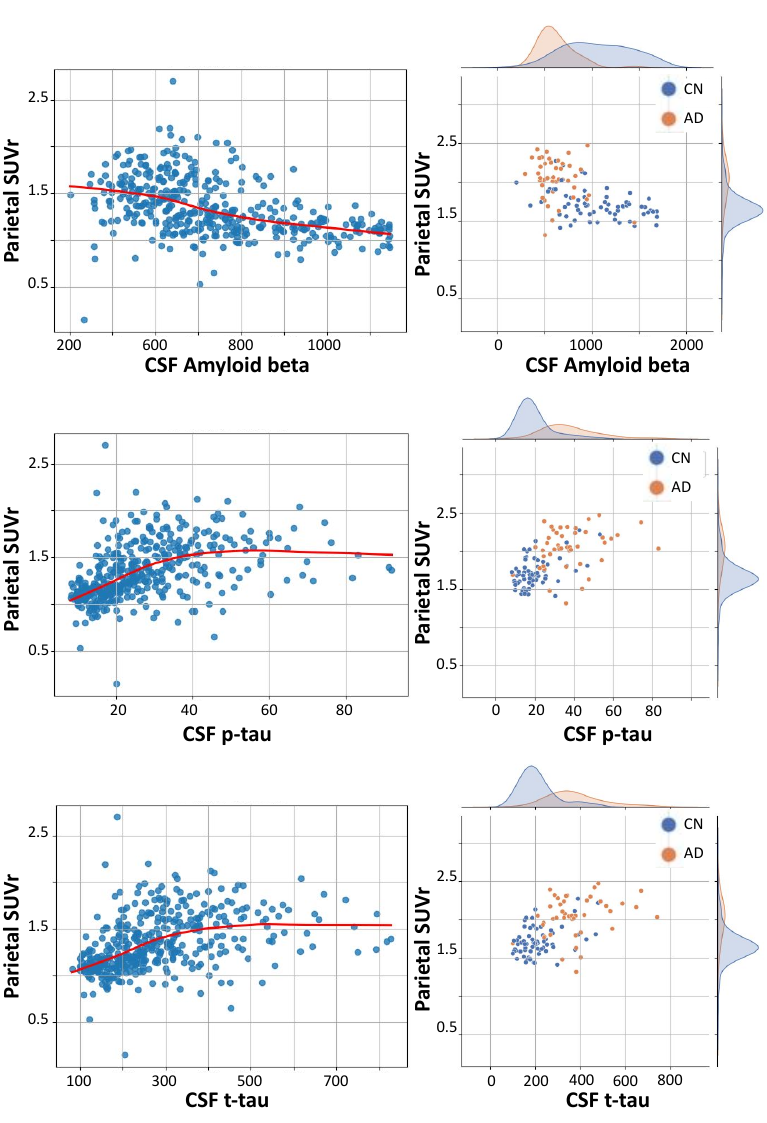}
    \caption{Relationship between PET standard uptake value for the parietal cortex and CSF biomarkers for the entire population (left column; $N=396$) and stable CN and AD subjects (right column; $N_{\text{CN}}=68$, $N_{\text{AD}}=45$). We overlay a local regression function to show the tendency of the relationship.}
    \label{fig:suvr-csf}
\end{figure}

Furthermore, we study the relationship between CSF and PET biomarkers over the AD continuum. We selected the parietal cortex SUVr as a representative of PET biomarker, but a similar behaviour can be found for the neocortex, temporal cortex, cingulate cortex and the DMN. In Figure~\ref{fig:suvr-csf} we show the relationship between amyloid SUVr and CSF amyloid beta, tau (t-tau) and phosphorylated tau (p-tau) over the AD continuum and the biomarker distribution for CN and AD subjects. We see an almost linear relationship between CSF amyloid beta and amyloid SUVr, whereas a non-linear tendency is detected between CSF t-/p-tau biomarkers and amyloid SUVrs. This latter result suggest that amyloid beta plateaus before t-tau and p-tau. Moreover, when restricting the analysis to CN/AD subjects we could distinguish between two clusters: one related to CN that has lower variability in both biomarkers than the other, whish is related to AD. Interestingly, CSF amyloid beta and amyloid SUVr appear to have complementary diagnostic value: while CSF amyloid beta is better suited to detect AD, amyloid SUVr seems better conditioned to discard AD due to the lower variability of CN.

\subsection{Multimodal analysis}
\begin{figure}[h]
    \centering
    \includegraphics[width=\columnwidth]{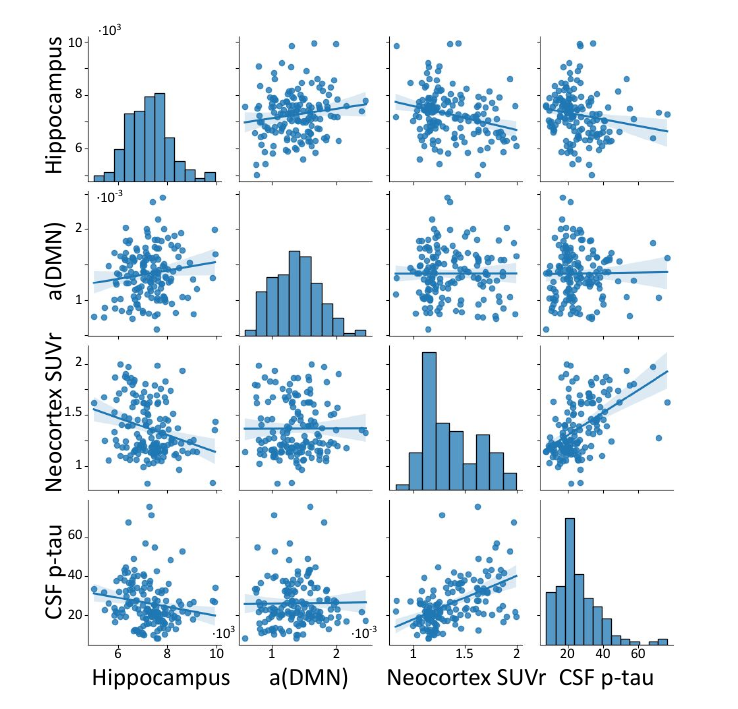}
    \caption{Pairwise relationship between different biomarkers available in the dataset ($N=152$). Concretely, we use CSF p-tau and image biomarkers derived from different modalities (T1w MRI, rs-fMRI and Amyloid PET). In the diagonal, we plot the distribution of each biomarker.}
    \label{fig:pairplot}
\end{figure}

One of the benefits of this pipeline is that we can easily compare multiple image modalities and CSF biomarkers. In Figure~\ref{fig:pairplot}, we show a combination of structural, functional, PET and CSF biomarkers across the continuum. We use the mean hippocampal value between hemispheres as a measure of neurodegeneration, which presents a mild correlation to the default model network activity (positive) and CSF p-tau (negative), while a stronger negative association with neocortex amyloid SUVr. The strongest association, as previously seen, is between CSF p-tau and amyloid SUVr in the neocortex. The default mode network amplitude appears to be unrelated to CSF and PET biomarkers

\section{Discussion and future work}
\label{sec:typestyle}
We have presented a registration method for multimodal brain imaging as well as a minimal preprocessing step to derive biomarkers readily to be used in multimodal analyses. This pipeline is suited for large-scale cohorts due to its robustness and fast inference times compared to standard pipelines.

One of the limitations of the pipeline is the PET-MR synthesis step. Here we use SynthSR, that has been optimised to work with multiple MRI contrasts and CT images but in practice it can also synthesise 1mm$^3$ T1w images from PET acquisitions. However, it struggles in the presence of noise, large PET uptake or low resolution acquisition. 

The extension of the method to other modalities (e.g., other MRI contrasts, tau PET) is straightforward. Moreover, we will incorporate this tool in our USLR framework presented in \cite{casamitjana2023USRL} for longitudinal and multimodal neuroimaging studies.


\vfill
\pagebreak

\section{Acknowledgments}
\label{sec:acknowledgments}
Adrià Casamitjana received funding from Ministry of Universities and Recovery, Transformation and Resilience Plan, through UPC (Grant No 2021UPC-MS-67573). R.S has received financial support from the Generalitat de Catalunya (2021-SGR00523), the María de Maeztu Unit of Excellence (Institute of Neurosciences, University of Barcelona, CEX2021-001159-M), and the Spanish Ministry of Science and Innovation (PID2020-118386RA-I00/AEI/10.13039/ 501100011033)

\bibliographystyle{IEEEbib}
\bibliography{strings,refs}

\end{document}